\documentclass[10pt,twocolumn,letterpaper]{article}

\usepackage{authblk}

\usepackage[pagenumbers]{cvpr} 

\usepackage{graphicx}
\usepackage{amsmath}
\usepackage{amssymb}
\usepackage{booktabs}
\usepackage{array}

\usepackage[dvipsnames]{xcolor}

\usepackage{pifont}

\usepackage{algorithm}
\usepackage{algpseudocode}
\usepackage{multicol}
\usepackage{balance}
\newcolumntype{C}[1]{>{\centering\arraybackslash}m{#1}}

\definecolor{cvprblue}{rgb}{0.21,0.49,0.74}
\usepackage[pagebackref,breaklinks,colorlinks,citecolor=cvprblue]{hyperref}

\def\paperID{} 
\def\confName{CVPR Workshop CV4Animals}
\def\confYear{2024}

\begin{document}

\title{Recurrence over Video Frames (RoVF) for the Re-identification of Meerkats}

\author[1,*]{Mitchell Rogers}
\author[1,*]{Kobe Knowles}
\author[1]{Gaël Gendron}
\author[1]{Shahrokh Heidari}
\author[1]{David Arturo Soriano Valdez}
\author[1]{Mihailo Azhar}
\author[1]{Padriac O’Leary}
\author[2]{Simon Eyre}
\author[1]{Michael Witbrock}
\author[1]{Patrice Delmas}

\affil[1]{NAOInstitute, The University of Auckland, Auckland, New Zealand}
\affil[2]{Wellington Zoo, Wellington, New Zealand}

\maketitle

\def\thefootnote{*}\footnotetext{These authors contributed equally to this work}\def\thefootnote{\arabic{footnote}}

\begin{abstract}
Deep learning approaches for animal re-identification have had a major impact on conservation, significantly reducing the time required for many downstream tasks, such as well-being monitoring. We propose a method called Recurrence over Video Frames (RoVF), which uses a recurrent head based on the Perceiver architecture to iteratively construct an embedding from a video clip. RoVF is trained using triplet loss based on the co-occurrence of individuals in the video frames, where the individual IDs are unavailable. We tested this method and various models based on the DINOv2 transformer architecture on a dataset of meerkats collected at the Wellington Zoo. Our method achieves a top-1 re-identification accuracy of $49\%$, which is higher than that of the best DINOv2 model ($42\%$). We found that the model can match observations of individuals where humans cannot, and our model (RoVF) performs better than the comparisons with minimal fine-tuning. In future work, we plan to improve these models by using pre-text tasks, apply them to animal behaviour classification, and perform a hyperparameter search to optimise the models further.
\end{abstract}

\section{Introduction}
In various areas of ecology and ethology, deep learning and computer vision have played significant roles in automating data collection by minimising the time required to extract information from images or videos \cite{Christin2019}. One such area is animal re-identification, which aims to match observations of the same individual across time or camera views \cite{Schneider2019}. Rapid methods for re-identification are crucial for processing the large-scale data required to make population estimates for many at-risk populations \cite{Tuia2022}, monitoring animal well-being in zoo environments \cite{Crockett2010}, and understanding animal behaviour using time budgets and enclosure usage maps \cite{Rose2021}. Traditional methods for these downstream tasks are typically laborious, require human observers to watch the enclosure for hours at a time, and are prone to observer errors \cite{Crockett2010}. 

Manual methods for animal re-identification are invasive and impractical in many cases. For meerkats (\textit{Suricata Suricatta}), previous methods include dye marks on fur \cite{Thornton_2012} or edible glitter to identify individuals by their faeces \cite{Scott_2014}. Tracking collars also allows researchers to record the behaviour of individuals without needing to re-identify individuals \cite{Chakravarty2019}. The visual appearance of meerkats varies, particularly in the colouration of the eye masks, around the snout, and body morphology such as tail length \cite{Heuls_2022}; however, these details are challenging for human observers to discern \cite{rogers2023meerkat}.

Patterns such as these on other animal species have enabled many recent deep learning methods for re-identification, such as dark markings on manta rays \cite{Moskvyak2020, Moskvyak2021}, stripes on Siberian tigers \cite{Li2020}, or ring patterns on seals \cite{Nepovinnykh2020, Nepovinnykh2022, Nepovinnykh2023}. One concern with these approaches is that focusing on animal-specific features results in methods that may not be applicable to other species \cite{Ravoor2020}. 

When the number of individuals is known in advance, re-identification can be treated as a classification problem \cite{Brookes2022,Andrew2020}. However, most animal re-identification applications have an open set of individuals, as populations in the wild and captivity change over time. Similarity-based re-identification aims to learn a similarity metric or distance function to compare the observations \cite{Schneider2020}. This allows individuals outside the training set to be recognised as well as previously seen individuals to be matched again \cite{Araujo2024}. 

Many previous re-identification methods used Convolutional Neural Networks (CNNs) trained using triplet loss to learn an embedding for each observation, which can be used to find matching individuals \cite{Araujo2024, Bouma2018}. Triplet loss networks learn using triplets of samples that include an anchor ($x_a$), a positive example of the same individual ($x_p$), and a negative example of a different individual ($x_n$) \cite{Hoffer2015, Schroff2015}. These networks aim to reduce the distance between observations of the same individual in an embedding space and increase the distance between different individuals. These models do not learn using trivial triplets, where the negative is clearly different from the anchor and positive observations. This has led to multiple variants of triplet loss being proposed that train the model using challenging triplets, such as the batch-hard triplet loss, in which for each anchor image, the furthest positive and closest negative within a batch of samples are selected \cite{Moskvyak2021,Li2020,Bouma2018,Hermans2017}. These triplet loss variants require the ground-truth labels of each individual to form challenging triplets, which is not always possible when domain experts cannot accurately distinguish between individuals.

In recent years, transformer models have exceeded the performance of many CNN models across several computer vision tasks \cite{ViT}. One of the most prominent transformer-based models is the DINO model \cite{DINO,DINOv2}, which uses a student-teacher training method to train a model to associate the global and local views of an image, leading to a greater semantic understanding of the scene and focus on foreground objects. However, these methods have yet to be applied to animal re-identification. Transformer-based models have also been successful in various video tasks \cite{ViViT}. Many re-identification studies using CNNs found that the performance can be improved using videos or sequences of frames rather than solely using images \cite{Zuerl2023, Nepovinnykh2023, Brookes2022}. This motivates our study, where the dataset consists of meerkats that may have identifying patterns in their motion and camouflage to their background. 

This study proposes a new similarity-based re-identification model, called Recurrence over Video Frames (RoVF), and a new variant triplet loss to re-identify animals in a zoo environment from videos in which ground-truth ID labels are unavailable. RoVF consists of a recurrent head on top of an image model that iteratively constructs a video embedding over frames. We instantiate the image model as DINOv2 \cite{DINOv2} with a Perceiver \cite{jaegle2021perceiver} as the recurrent architecture. We compare this architecture to various pretrained and fine-tuned DINO models~\cite{DINO} and achieve a better top-1 accuracy. We evaluate all models on a meerkat dataset collected at the Wellington Zoo (Wellington, New Zealand) \cite{rogers2023meerkat} and achieve viable meerkat re-identification where human annotators could not.

\section{Methodology}

\subsection{Dataset creation}
\label{sec:dataset-creation}
A new re-identification dataset was generated based on the twenty 12-minute videos provided in the Meerkat Behaviour Recognition Dataset \cite{rogers2023meerkat}. The meerkats were labelled as tracks without a fixed ID per meerkat, meaning that if a meerkat exited the camera view and returned, it was assigned a new ID by the annotators since the annotators could not distinguish individuals. The unoccluded tracks of individual meerkats were divided into short 10-second clips. These clips were cropped from the full-resolution videos by determining the maximum dimensions of the meerkats bounding box over the duration of the clip and creating a fixed-sized square centred on the meerkats bounding box for the duration of the clip. These crops were then resized to a resolution of 224 × 224 pixels at 1 fps. To generate more data, the start of the 10-second clips was stagnated once every 3$\frac{1}{3}$ seconds. 

Similar to other animal re-identification datasets, this dataset presents challenges. Meerkats are non-rigid and deformable, leading to a high variation in posture \cite{Li2020,Nepovinnykh2022}. The zoo enclosure also emulates a natural environment, leading to variations in lighting, occlusions, and meerkats camouflaging with their surroundings. Another challenge of this dataset is the size of the meerkats; their bounding boxes are often less than 100 pixels in width or height, making the patterns of their fur difficult to see. To deal with low-resolution observations, clips were discarded if the bounding box of the meerkat did not exceed 70 pixels.

Clips were extracted from 16 of the 20 full videos provided in the full dataset for training, and clips from the other four videos were held for testing, resulting in 8,244 clips in the training set and 1,800 in the test set, respectively. Across the 20 videos, 1087 unique IDs were observed, averaging 54 unique IDs per video, despite there being only 15 adult meerkats in the enclosure during the recordings. Of these IDs, 93 in the training set and 18 in the test set had less than three clips assigned to them because these meerkats were not in view for long during the videos. These were removed from being selected as positive/anchor clips, but could be selected as negative examples.

\subsection{Triplet loss}
\label{sec:triplet-loss}
The triplet loss method requires training data in the form of triplets with one anchor, one observation from the same meerkat (positive), and one observation from another meerkat (negative). Two issues arise from this dataset: individuals can change IDs, and individual tracks have limited background variation. The annotators could not re-identify each meerkat when they reentered the scene; therefore, negative pairs must be sampled from tracks that share at least one frame to eliminate the possibility of the same meerkat being used as all three observations in a triplet. Owing to the limited background variation for each annotation track, the trained models may rely on background information, such as sand or bark, in some parts of the enclosure to match individuals.

\begin{figure}[ht]
    \centering
    \includegraphics[trim={2cm 1.35cm 1.6cm 0.6cm},clip,width=0.98\linewidth]{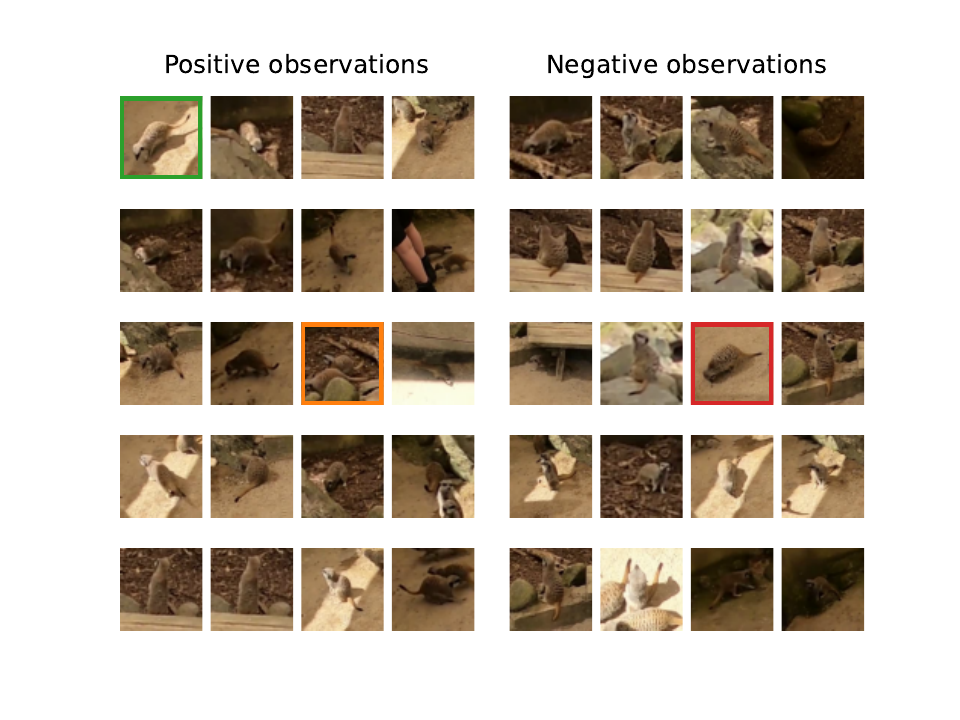}
    \caption{Example positive and negative sets. The first frame of 20 positive clips of the same meerkat (left half) and 20 negative clips of other meerkats (right half). The anchor (green), positive (orange), and negative (red) have been selected based on embeddings from the training ResNet model.}
    \label{fig:Triplets}
\end{figure}

Using an approach similar to the \textit{batch-hard} triplet mining method \cite{Hermans2017}, we propose a hard triplet mining strategy to reduce the number of trivial triplets when individual IDs are unavailable. Triplets are first generated by randomly selecting an anchor individual from all clips and then randomly selecting up to $j$ positive examples ($P$) of that individual and up to $k$ negative examples ($N$) from other meerkats observed in the same frame. In this study, we set $j=k=20$, meaning that up to 20 positive and negative clips were selected for each evaluation. Fewer than 20 clips are available for some tracks, where the meerkats do not remain in the video frame for long. During training, the embeddings of all clips are computed, and the most difficult pair of positive clips and the most difficult negative clip were selected for training. The most difficult positive pair was defined as the two clips with the largest embedding distance from each other in the positive set, and the most difficult negative observation was defined as the observation closest to one of the positive clips. The closest positive clip then becomes the anchor. Figure \ref{fig:Triplets} shows an example of positive and negative observations randomly selected for one individual. The anchor, positive and negative, selected for training by an image-based model are highlighted.

\subsection{Model architecture}
\label{sec:model-architecture}

\begin{figure}
    \centering
    \includegraphics[width=\linewidth]{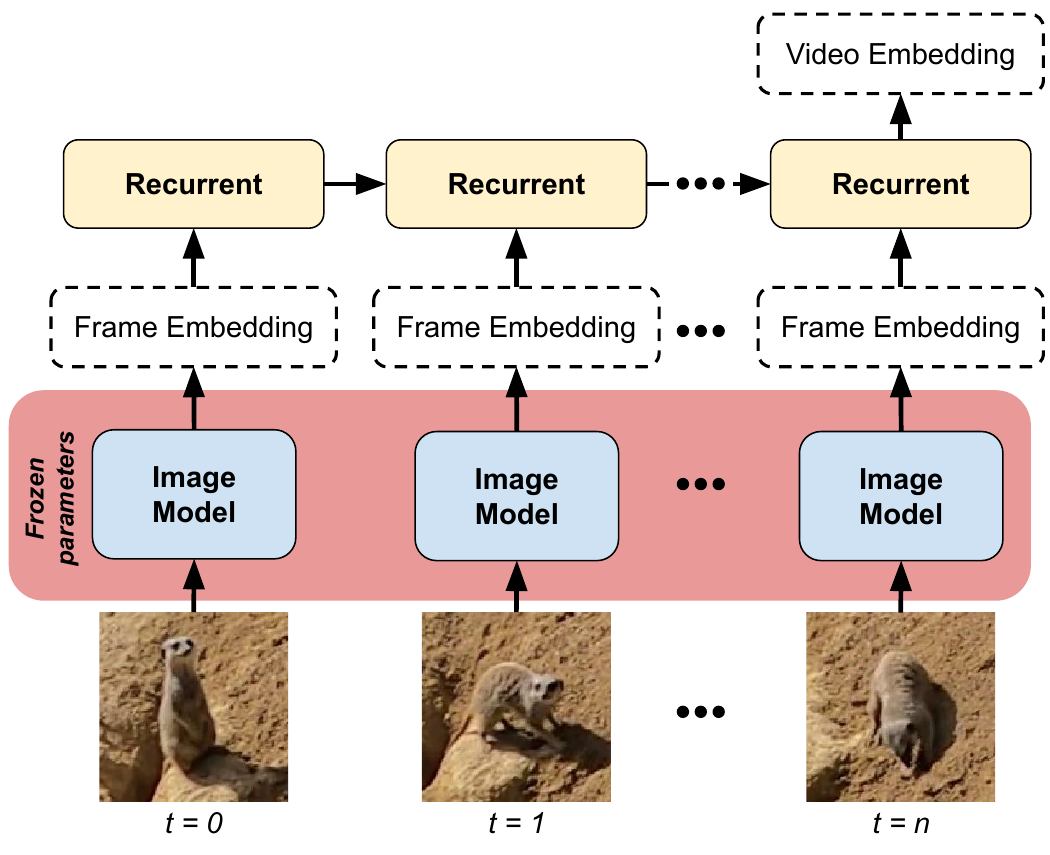}
    \caption{\textit{Recurrence over Video Frames} (RoVF) is an architecture that adds a recurrent component on top of an existing image model---that outputs image/frame embeddings---allowing representations over a video to be constructed. The Recurrent architecture iteratively, over frames, builds a representation of the video from the image model's embeddings for a frame; after the last frame, a video embedding is outputted by the recurrent model.}
    \label{fig:dino-recurrent-architecture}
\end{figure}

\begin{algorithm}
\caption{Recurrence over Video Frames (RoVF)}
\label{alg:RoVF}
\begin{algorithmic}[1]
\Require Video $V \in \mathbb{R}^{f \times c \times w \times h}$, recurrent model $R$ with parameters $\theta_r$ and hidden state $r_h \in \mathbb{R}^d$, and an image model $I$ with parameters $\theta_i$
\Procedure{RoVF}{$V$}
    \For{$i \gets 1$ \textbf{to} $f$}
        \State $f_{emb}^i \gets I(V[i,:,:,:])$ \Comment{process the $i^{\text{th}}$ frame}
        \State $r_h, v_{emb} \gets R(r_h, f_{emb}^i)$ \Comment{output the next recurrent hidden state and video embedding prediction}
    \EndFor
    \State \textbf{return} $v_{emb}$ \Comment{return the video embedding after the last frame is processed}
\EndProcedure
\end{algorithmic}
\end{algorithm}

For the problem of re-identification with video, we introduce an architecture---depicted in Figure \ref{fig:dino-recurrent-architecture}---called \textit{Recurrence over Video Frames} (RoVF) that utilises a recurrent architecture on top of an existing pre-trained image model. In our experiments the recurrent architecture is a Perceiver transformer \cite{jaegle2021perceiver} and the image model is a DINOv2 variant \cite{DINOv2}. 

Algorithm \ref{alg:RoVF} shows how a video embedding is constructed from a video of $f$ frames, $c$ channels, of width $w$, and of height $h$, by RoVF. Each frame of the video $V[i,:,:,:]$ is processed by the image model (line 3), where $i$ represents the $i^{\text{th}}$ frame. The frame embedding $f_{emb}^i$ for the $i^{\text{th}}$ frame is input to the recurrent architecture, with the previous hidden state $r_h$, which produces a new hidden state $r_h$ and a video embedding $v_{emb}$ (line 4). This is repeated for each frame (lines 2--4), with the last video embedding $v_{emb}$ being returned by the algorithm (line 6). When using the Perceiver architecture, the hidden state is referred to as the latent array.

For further details on the recurrent architecture used in this paper, see Perceiver~\cite{jaegle2021perceiver}. For the image model, see DINO~\cite{DINO} and DINOv2 ~\cite{DINOv2}.




\subsection{Evaluation metrics}
\label{sec:evaluation-metrics}
The most common approach for evaluating animal re-identification methods is to split the test set into query images and a gallery set, where gallery observations are ranked in terms of the distance from the query image to find possible matches \cite{Bouma2018,Ravoor2020}. The top-k accuracy metric then measures how often the query image has the same ID as one of the k highest-ranked gallery images. This dataset does not have ID labels for each individual; therefore, the top-k metrics must be calculated based on the clips of individuals who have frames in common. 

The test set contained 87 individual tracks, and sets of two positives and nine negatives from clips of meerkats occurring in the same frames were randomly generated. These sets were then manually screened to remove sets where multiple individuals were huddling and sets where the identification of the positive pair was visually trivial, such as when the positive pair had a background distinctly different from the negative examples. In total, 100 sets were generated using this method. For each set, the positive examples were alternated as the query clip, with the remaining non-query clips used as the gallery of clips, resulting in 200 testing sets of query and gallery clips. Embeddings were extracted using each re-identification architecture, and the proportion of sets in which the correct match was the closest embedding (in Euclidean space) to the query (top-1) and the proportion where the correct match was in the closest three gallery examples (top-3) was recorded.

\subsection{Experiments}




We instantiate one version of RoVF, outlined in Section \ref{sec:model-architecture}, with DINOv2 base \cite{DINOv2} as the image model and the Perceiver \cite{jaegle2021perceiver} as the recurrent model (RoVF-Base). The Perceiver model is set up with a latent array, input dimension, latent dimension, and output embedding size of 768; 2 transformer layers; dropout rate of 0.1. Dropout is applied after every frame embedding and in the default positions of PyTorch's TransformerEncoderLayer.

As baselines, we include four pre-trained DINOv2 variants (small, base, large, and giant) and average the patch embeddings in each frame and then all frame embeddings, resulting in a video embedding. Additionally, we fine-tune the small and base variants of DINOv2 with the same averaging regime. 

We used the ResNet-50 model \cite{he2015deep} pretrained on the ImageNet dataset to compare with a typical image-based animal re-identification model. ResNet models have been used in many other studies for animal re-identification and commonly exceed the performance of other common feature extraction methods \cite{Borlinghaus2023, Bouma2018, Moskvyak2020, Schneider2020}. Images (224 $\times$ 224) were obtained from the first frame of each clip and trained using triplets generated using the same hard-triplet mining method. We fixed the weights of the model and exclusively trained the fully connected layer with a dimensionality of 2048 to derive an embedding vector of 256 dimensions. To compare this with video-based models, embeddings were calculated from the first frame of each video in the test set and evaluated in the same manner as the video models.

For all experiments reported, we use a learning rate schedule with a linear warmup from 0.0001 to 0.0005 over 5\% of an epoch; cosine decay down to 0.00001 over the duration of training. All models are trained on a single NVIDIA A100 80GB GPU. A batch size of 30 is used for all models; each batch consists of 10 anchor, positive, and negative pairs---an effective batch size of 10. We use PyTorch's TripletMarginLoss with default hyperparameters (no difference was found when modifying the margin value). Given that we only have a test set (no validation set), all video-based models' results are reported for the $5^{\text{th}}$ or $10^{\text{th}}$ epochs, and the ResNet-50 model is reported for the $15^{\text{th}}$ epoch.


\section{Results}

\begin{table}[!ht]
\centering
\resizebox{\linewidth}{!}{
\begin{tabular}{m{2.4cm}C{0.8cm}C{0.9cm}C{0.9cm}C{0.9cm}C{0.9cm}}
\hline
Model/size & Fine-tuned & Epochs & Top-1 & Top-3 & Avg. Epoch Time \\
\hline \hline
Random          &  -  & - & 10.0 & 30.0 & - \\
ResNet-50        & \checkmark    & 15 & 26.5 & 52.5 & 0.17 \\
DINOv2-Small    & \ding{55}     & - & 41.5 & 72.5 & - \\
DINOv2-Base     & \ding{55}     & - & 38.5 & 73.0 & - \\
DINOv2-Large    & \ding{55}     & - & 42.0 & \textbf{73.5} & - \\
DINOv2-Giant    & \ding{55}     & - & 42.0 & \textbf{73.5} & - \\
DINOv2-Small    & \checkmark    & 5 & 26.0 & 63.5 & 4.3 \\ 
DINOv2-Small    & \checkmark    & 10 & 33.5 & 63.5 & 4.3 \\ 
DINOv2-Base     & \checkmark    & 5  & 30.0 & 64.5 & 5.8 \\ 
DINOv2-Base     & \checkmark    & 10 & 29.0 & 62.0 & 5.8 \\ 
RoVF-Base       & \ding{55}     & 5  & 48.0 & 63.0 & 7.0 \\ 
RoVF-Base       & \ding{55}     & 10 & \textbf{49.0} & 64.5 & 7.0 \\
\hline
\end{tabular}}
\caption{Test set performance for each model and variant. ``Fine-tuned'' refers to if the image model of each model is fine-tuned (for RoVF, the recurrent architecture is always fine-tuned). ``Epochs'' shows how many training epochs for the reported results. ``Avg. Epoch Time'' is the average time in hours for a training epoch.}
\label{table:test-set-results}
\end{table}

\begin{figure*}[htb]
    \centering
    \includegraphics[width=0.98\linewidth]{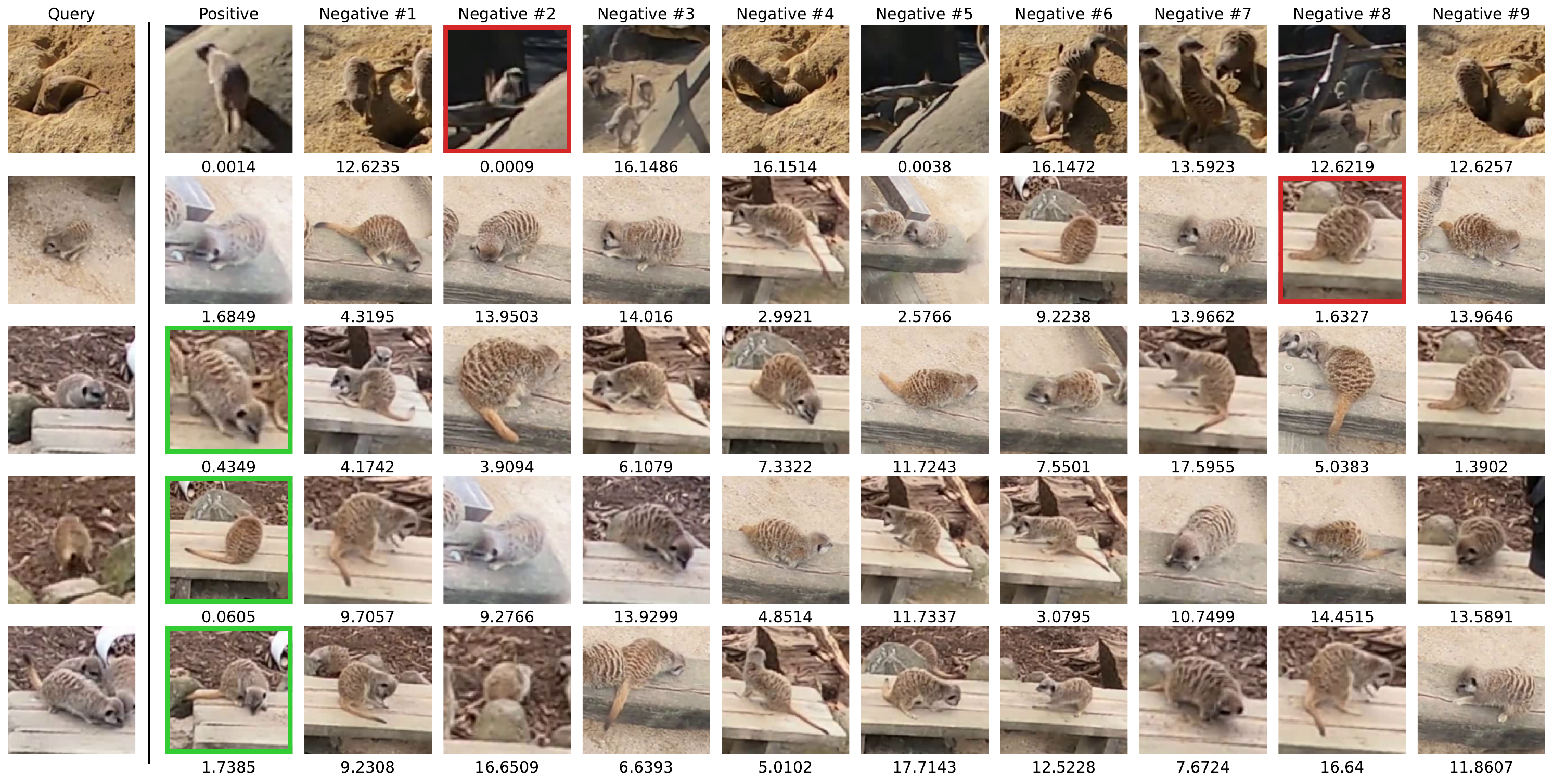}
    \caption{Examples of incorrect (red) and correct (green) re-identifications of a query clip (left-most column) using the best RoVF model. The embedding distance between the query and gallery clip is shown underneath each thumbnail.}
    \label{fig:Examples}
\end{figure*}

Table \ref{table:test-set-results} showcases the results of our experiments. We observe that all DINOv2 variants with no fine-tuning (rows 3--6) show good baseline performance by averaging the embeddings over patches and then frames. No success was found when fine-tuning these models on our dataset (rows 7--10); performance degraded compared to the pre-trained models. RoVF-Base achieved the best top-1 accuracy over all models (final two rows), demonstrating the potential of this architecture; however, the top-3 accuracy degraded. Unsurprisingly, RoVF-Base has the longest average epoch running time, with the Perceiver's addition adding 1.2 hours to the average epoch time. Most video-based models outperformed the image baseline, ResNet-50 considerably. For this application, video information may offer additional insights into individuals through their motion over consecutive frames. Because meerkats remain stationary for an extended period and the tracks are limited, meerkats in randomly sampled frames from the same track are likely to share similar poses and backgrounds. A human observer may look at the tail or eye mask of the meerkat that may be occluded in a randomly selected frame.



It is interesting to highlight that fine-tuning DINOv2 degraded the performance, but freezing its parameters and adding a recurrent head (RoVF) improved the performance over the pre-trained DINOv2 models. This may be due to the choice of averaging over patches and then frames for DINOv2 instead of concatenating embeddings across patches and then averaging the embeddings. Another reason could be that there is an insufficient amount of training data or a lack of diversity, which gives RoVF an advantage. In addition, although a limited search for good hyperparameters has been conducted, an extensive hyperparameter search may result in better performance in some instances. Another concern is trivially simple test cases. Although care was taken when constructing the test examples, these models may rely on background cues rather than the features of the individual, where the positive example is visually similar to the anchor. We would like to mitigate overfitting to the background in future work with pre-text training and data augmentation.

Figure \ref{fig:Examples} shows five test cases of the ROVF model applied to match the query clip with one of ten gallery clips. These examples demonstrate that the model can distinguish between meerkats in similar environments, and for the two negative examples, the distance of the query is close to the selected match. However, because of the nature of the dataset, we cannot see how well the model performs when matching meerkats observed in vastly different backgrounds. 


\section{Conclusion}
This study is an initial proof-of-concept of a transformer-based architecture for animal re-identification without individual labels, called Recurrence over Video Frames (RoVF). Our models achieved good results compared to various models based on the DINOv2 architecture when applied to a dataset of short clips of meerkats, with a top-1 re-identification accuracy of 49\%. Given that the original annotators of the dataset were unable to identify individuals better than random chance, these results were impressive. In future work, we aim to extend this method to video behaviour classification with behaviour labels, compare our results to those of other methods (frame- or video-based), and improve the results using pre-text training. We also intend to perform a more rigorous hyperparameter search to find a more optimal number of training epochs and FPS.


\onecolumn
\begin{multicols}{2}
{
\section*{Ethics statement}
Ethical approval was waived for this study due to the camera-based and non-invasive approach towards data collection and re-identification. No manipulation of the meerkats, collars or markings were used for re-identification. Data acquisition was exclusively carried out at Wellington Zoo during standard opening times, without any changes to the management procedures of the animals.
\small
\bibliographystyle{ieeenat_fullname}
\bibliography{egbib}
\balance
}
\end{multicols}

\end{document}